\documentclass[10pt,twocolumn,letterpaper]{article}

\usepackage{cvpr}              

\usepackage{graphicx}
\usepackage{amsmath}
\usepackage{amssymb}
\usepackage{booktabs}
\usepackage{float}
\usepackage{multirow}
\usepackage[accsupp]{axessibility}

\usepackage[pagebackref,breaklinks,colorlinks]{hyperref}

\usepackage[capitalize]{cleveref}
\crefname{section}{Sec.}{Secs.}
\Crefname{section}{Section}{Sections}
\Crefname{table}{Table}{Tables}
\crefname{table}{Tab.}{Tabs.}

\def\cvprPaperID{02884} 
\def\confName{CVPR}
\def\confYear{2022}

\begin{document}

\title{Towards Robust Rain Removal Against Adversarial Attacks: A Comprehensive Benchmark Analysis and Beyond}

\author{Yi Yu$^{1,2}$ ~\quad \quad Wenhan Yang$^{1}$\thanks{Corresponding author.} \quad\quad ~Yap-Peng Tan$^1$ \quad\quad~ Alex C. Kot$^1$ \quad\quad
\\
$^1$School of Electrical and Electronic Engineering, Nanyang Technological University\\
$^2$ROSE Lab, Interdisciplinary Graduate Programme, Nanyang Technological University\\
{\tt\small yuyi0010@e.ntu.edu.sg ~~\quad \{wenhan.yang,eyptan,eackot\}@ntu.edu.sg}
}
\maketitle

\begin{abstract}
Rain removal aims to remove rain streaks from images/videos and reduce the disruptive effects caused by rain.
It not only enhances image/video visibility but also allows many computer vision algorithms to function properly.
This paper makes the first attempt to conduct a comprehensive study on the robustness of deep learning-based rain removal methods against adversarial	attacks.
Our study shows that, when the image/video is highly degraded, rain removal methods are more vulnerable to the adversarial attacks as small distortions/perturbations become less noticeable or detectable.
In this paper, we first present a comprehensive empirical evaluation of various methods at different levels of attacks and with various losses/targets to generate the perturbations from the perspective of human perception and machine analysis tasks.
A systematic evaluation of key modules in existing methods is performed in terms of their robustness against adversarial attacks.
From the insights of our analysis, we construct a more robust deraining method by integrating these effective modules.
Finally, we examine various types of adversarial attacks that are specific to deraining problems and their effects on both human and machine vision tasks, including 1) rain region attacks, adding perturbations only in the rain regions to make the perturbations in the attacked rain images less visible; 2) object-sensitive attacks, adding perturbations only in regions near the given objects. {Code is available at {\url{https://github.com/yuyi-sd/Robust_Rain_Removal}}.}
\end{abstract}

\section{Introduction}
Rain removal methods aim to remove the disruptive effects caused by rain streaks to restore a clean version of the image.
It not only largely improves the visibility of the rainy image but can also improve the performance of many subsequent (downstream) computer vision applications.

Early approaches are mainly model-driven and address the deraining problem based on the statistical properties of rain streaks and background scenes, \textit{e.g.} image decomposition~\cite{ID}, sparse coding~\cite{DSC}, and Gaussian mixture model~\cite{LP}.
These methods can well handle light rain.
However, they fall short in cases of handle heavy rain and often blur the background scenes.
Recently, deep-learning based deraining methods~\cite{DBLP:journals/pami/YangTFGYL20,Deep_detail_net} have become the mainstream.
These methods have the capacity to model more complicated mappings from rain images to clean images and offer a better performance in terms of less remaining rain streaks and better preserved background scene.
They also enhanced the performance of practical applications such as video surveillance.

\begin{figure}[t]
\centering
\includegraphics[scale=0.23]{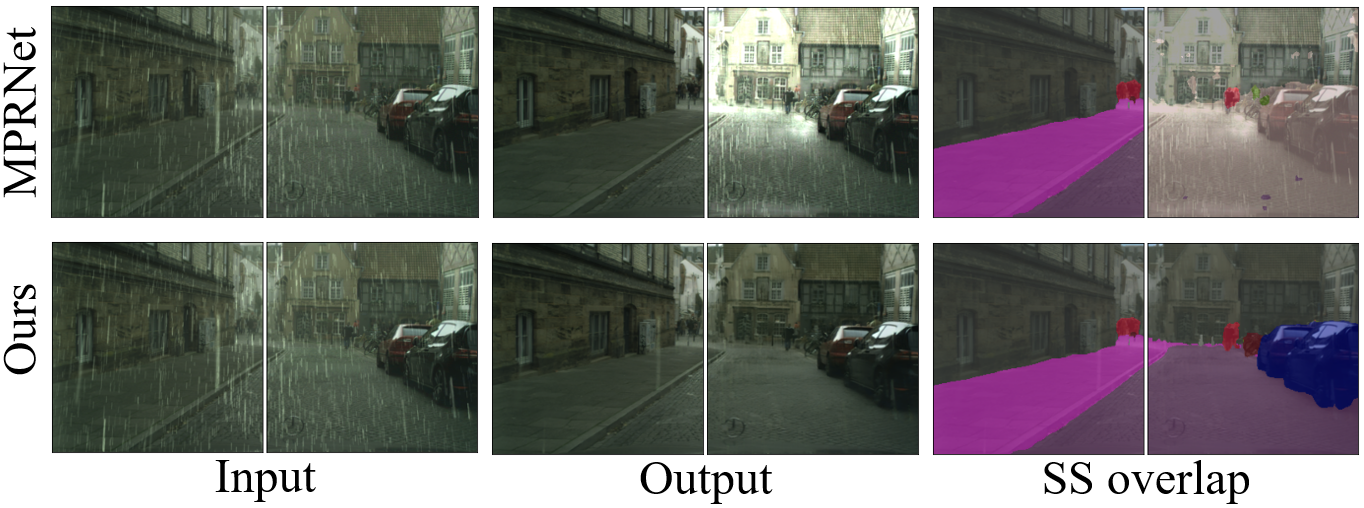}
\vspace{-7mm}
\caption{Left is from corresponding patch of clean input/output, and right is from adversarial input/output with perturbation bound $\epsilon=4/255$. The semantic segmentation overlap of output is shown in the last column. The testing image is from RainCityscape~\cite{DBLP:conf/iccv/HalderLC19}.
}
\vspace{-5mm}
\label{figure0}
\end{figure}

While many deep learning-based deraining methods have been introduced, it isn't a thorough study on the robustness of these methods against adversarial attacks.
It is also a real concern of deraining methods that some unnoticeable perturbations can alter the predicted results of deep networks~\cite{Robustness1,Robustness2}.
As the rainy images can be usually highly degraded by intensive rain streaks, perturbations can be easily and noticeably hidden in such images.
These adversarially generated outputs can also compromise the reliability and stability of the subsequent applications such as video surveillance and autonomous driving, that adopt rain removal methods as a pre-processing module.

In this paper, we make the first attempt to investigate, improve, and evaluate the robustness of deep learning-based rain removal methods against adversarial attacks. Our main contributions are summarized below.
\begin{itemize}
	\item 
	A thorough analysis of existing deraining methods at different levels of robustness against adversarial attacks and with various losses/targets to generate the perturbations, from both the perspectives of human and machine vision tasks.
	\item A systematic evaluation of the modules in existing methods at both the model side  (recurrence, attention, receptive field, and side information) and loss side (adversarial loss) in terms of their impact on robustness.
	\item A more robust deraining method constructed by integrating the effective modules which are identified from our analysis.
	\item Investigation of various adversarial attack types and scenarios specific to rain removal degradation in terms of both human and machine vision tasks. 
\end{itemize}

\section{Related Work}
\subsection{Adversarial Attacks}
Deep neural networks~\cite{intriguing} are vulnerable to adversarial examples that are visually indistinguishable from natural images but can compromise the performance of models.
From the attack method aspect, Szegedy \textit{et al.} \cite{intriguing} developed an optimization-based approach to find adversarial examples within specific amounts of input perturbation.
Goodfellow \textit{et al.} \cite{Robustness2} proposed the Fast Gradient Sign Method (FGSM) and Kurakin \textit{et al.} \cite{DBLP:conf/iclr/KurakinGB17} developed multiple variations.
Madry \textit{et al.} \cite{DBLP:conf/iclr/MadryMSTV18} proposed a more powerful multi-step variant, which is essentially Projected Gradient Descent (PGD). 
There are also studies on the effects of adversarial attacks on various computer vision applications and scenarios, \textit{e.g.} human pose estimation~\cite{NIPS2017_d494020f}, object detection~\cite{Xie_2017_ICCV}, and semantic segmentation~\cite{Xie_2017_ICCV}.

There are also works on adversarial attacks and robustness analysis in low-level processing tasks.
In~\cite{defense}, super-resolution is adopted as the tool to mitigate the adversarial attacks for deep image classification models.
In~\cite{deep_fool}, adversarial attacks are exploited on super-resolution model to attack the down-stream computer vision tasks, \textit{e.g.} image classification, style transfer, and image captioning.
In~\cite{sr_robustness}, the robustness of deep learning-based super-resolution methods against adversarial attacks is examined.
In~\cite{2020arXiv200909205Z,2021arXiv210413673G}, rain streak/haze is considered as one type of adversarial noise.
In~\cite{2021arXiv210415022C}, the robustness of deep image-to-image models against adversarial attacks is evaluated from the perspective of human vision.
However, there is no work investigating adversarial attacks and robustness analysis in rain removal scenarios.
It is also absent to consider the adversarial attacks from the perspective of human and
machine vision tasks.

\subsection{Single-image rain removal}
It is non-trivial to separate rain streaks and background scenes from their mixed versions.
Methods based on signal separation and texture classification have been proposed to address the challenge. Early non-deep learning-based works~\cite{ID,chen2013generalized,nonlocal_derain,DSC,LP} are constructed based on statistical models.
In~\cite{ID}, an attempt is made to perform the single-image deraining based on image decomposition using morphological component analysis.
In a successive work~\cite{DSC}, a mutual exclusivity property is incorporated into a sparsity model to form a discriminative sparse coding that can better separate the rain/background layers from their non-linear composite.

Recently, deep-learning based deraining methods~\cite{Deep_detail_net,DBLP:journals/pami/YangTFGYL20,2017arXiv171206830L,Zhang_2018_CVPR} have been proposed to obtain superior rain removal results.
More recent works~\cite{NLEDN, DBLP:conf/cvpr/ZamirA0HK0021,Wang_2019_CVPR,lightweight,all_in_one} also focus on advanced deep networks.
Others~\cite{MSPFNet, DBLP:conf/cvpr/YasarlaP19, heavy_rain, DBLP:conf/cvpr/WangX0M20, detail_recovery,con_var,yang2020self,yang2020aaai,yang2019visibility} make use of effective priors for deraining.
Some studies focus on the generalization capability of deraining models. In~\cite{semi_supervised}, Wei \textit{et al.} make use of real rainy images in network training by formulating the residual between an input rainy image and its expected deraining result as a mixture of Gaussians.
In~\cite{Syn2Real}, a Gaussian process-based semi-supervised learning is proposed to allow the network learn with synthetic data and generalize deraining performance with unlabeled real rainy images. These methods can provide better deraining results in terms of less remaining rain streaks and rich background textures. However, there is little work on the vulnerability of these deep networks against adversarial attacks. This work aims to fill this gap.

\section{Benchmarking Adversarial Robustness of Deraining Models}
\subsection{Attack Framework}
\label{sec:attack_framework}
Adversarial attacks aim to deteriorate the output of the deraining methods by adding a small amount of visually unperceivable perturbations to the input rainy images.
To generate the adversarial perturbations, we develop an attack method based on one optimization-based method PGD \cite{DBLP:conf/iclr/MadryMSTV18}, which has been extensively used and considered as a powerful attack method so far for evaluating the robustness of classification models\cite{DBLP:conf/icml/Croce020a}.

We consider a deraining model $f\left(\cdot | {\theta}\right)$ parametrized by $\theta$.
We denote $X$ as the input image, $Y$ as the groundtruth image,  $\delta$ as the perturbations, and $D$ as the metric to measure the degradation. 
The objective of adversarial attacks is to maximize the deviation of the output from the target, \textit{e.g.} the original output:
\begin{equation}
    \delta = \underset{\delta, {\left\|\delta\right\|}_p \leq \epsilon}{\arg\max} D\left( f(X|\theta), f(X+\delta |\theta))\right).
\end{equation}
To solve the maximization problem with $\ell_p$-norm bound constraint (usually $\ell_{\infty}$-norm), we use the PGD approach to obtain the perturbations iteratively:
\begin{align}
    {\omega}^{t+1}&=
    {\delta}^{t}+\alpha \text{sgn}
    \left(\nabla_{\delta}
    D\left( f(X|\theta), f(X+\delta |\theta))\right)\right), \label{eq:1}\\
    {\delta}^{t+1}&=\text{clip}_{\left[-\epsilon,\epsilon\right]\cap\left[-X,1-X\right]}({\omega}^{t+1}),\label{eq:2}
\end{align}
where $\nabla$ represents the gradient operation, $\text{sgn}$ extracts the sign of gradients, and the $\text{clip}$ operation guarantees that the perturbations are within $\left[-\epsilon,\epsilon\right]$ and the perturbed input is within $\left[0,1\right]$. The term $\alpha$ controls the step length each iteration, and $\epsilon$ represents the maximum perturbation allowed for each pixel value. The initial $\delta^0$ is sampled from the uniform distribution $U{(-\epsilon,\epsilon)}$, and the final adversarial perturbations $\delta^{T}$ is obtained after $T$ iterations.

Based on different attack objectives, we define two types of corresponding metrics for $D$:
\begin{itemize}
    \item \textit{Restoration} (\textit{Human Vision}): per-pixel measures, \textit{e.g.} $\ell_2$ Euclidean distance:
    \begin{equation}
    \delta = \underset{\delta, {\left\|\delta\right\|}_{\infty} \leq \epsilon}{\arg\max} \left\|f(X+\delta|\theta)-f(X|\theta)\right\|_2.
    \end{equation}
    \item \textit{Downstream CV Tasks} (\textit{Machine Vision}): the distance of feature extracted from pretrained models, \textit{e.g.} Learned Perceptual Image Patch Similarity (LPIPS)~\cite{zhang2018perceptual}:
    \begin{equation}
    \delta = \underset{\delta, {\left\|\delta\right\|}_{\infty} \leq \epsilon}{\arg\max} \ell_{pips}(f(X+\delta|\theta),f(X|\theta)).
\end{equation}
\end{itemize}
We denote the first one as LMSE attack and the second as LPIPS attack.
For each attack type and perturbation bound, $\delta_{D,\epsilon}(X)$ is obtained as Eqs. (\ref{eq:1}--\ref{eq:2}), and the general form of adversarial robustness
for deraining model $f$ is given by
\begin{align}
    R_f^{D,\epsilon} =  \mathbb{E}_{X \sim \mathbb{P}_{data}}\left[ P(Y,f(X+\delta_{D,\epsilon}(X))) \right],
\end{align}
where the samples follow the distribution $\mathbb{P}_{data}$, and $P$ is the evaluation metric. While it is impossible to evalaute the robustness by considering all values of $\epsilon$, we estimate it using finite $\epsilon$ and limited test samples. With the estimation, we use mean-Adversarial-Performance (mAP) to evaluate the adversarial robustness against each type of attack $D$:
\begin{align}
    mAP_{\scriptscriptstyle D} = \frac{1}{n(E)}\sum_{\epsilon \in E}\frac{1}{n(\mathcal{D}_t)}\sum_{X \in \mathcal{D}_t}P(Y,f(X+\delta_{D,\epsilon}(X))),
\end{align}
where $E$ is the set of $\epsilon$ to be evaluated, $\mathcal{D}_t$ is the test dataset, and $n(\cdot)$ counts the numbers. Note that $P$ is usually measured in terms of PSNR and SSIM.

\subsection{Evaluation Metrics}
In our benchmark, we consider two types of performance evaluation metrics:
\begin{itemize}
    \vspace{-1mm}
    \item The commonly used quality measures including peak signal to noise ratio (PSNR) and structural index similarity (SSIM)~\cite{DBLP:journals/tip/WangBSS04}, for evaluating signal fidelity. 
    \vspace{-1mm}
    \item Task-driven metrics that evaluate the performance of down-stream machine analysis tasks.
    Specifically, we evaluate the performance of object detection and semantic segmentation, \textit{i.e.}  mean Intersection over Union  (mIoU) for semantic segmentation and Average Precision (AP) for pedestrian detection, with the approaches SSeg~\cite{DBLP:conf/cvpr/0001SRSNTC19} and Pedestron~\cite{Hasan_2021_CVPR} on the RainCityscape dataset~\cite{DBLP:conf/iccv/HalderLC19}.
\end{itemize}

\begin{table}[t]
    \small
    \centering
    \caption{PSNR, parameter and module comparisons on two datasets. AT denotes attention module, RB recurrent blocks, SI side information and DD diverse dilations, and RC RainCityscape. }
    \vspace{-3mm}
    \begin{tabular}{@{\hspace{1.1pt}}c@{\hspace{1.1pt}}||@{\hspace{2pt}}c@{\hspace{2pt}} ||@{\hspace{2.2pt}}c@{\hspace{2.2pt}}|@{\hspace{2pt}}c@{\hspace{2pt}}|@{\hspace{4.0pt}}c@{\hspace{4.0pt}}|@{\hspace{1pt}}c@{\hspace{1pt}}||@{\hspace{1.1pt}}c@{\hspace{1.1pt}}|@{\hspace{3pt}}c@{\hspace{3pt}}}
     \hline
     Model & Para. & AT& RB & SI & DD & Rain100H & RC \\
     \hline
     JORDER-E~\cite{DBLP:journals/pami/YangTFGYL20}&4.17M&&\checkmark&\checkmark&\checkmark&29.75&32.51\\
     RCDNet~\cite{DBLP:conf/cvpr/WangX0M20}&3.17M&&\checkmark&&&29.65&31.44 \\
     MPRNet~\cite{DBLP:conf/cvpr/ZamirA0HK0021}&3.64M&\checkmark&&&&30.56&36.31\\
     PReNet~\cite{DBLP:conf/cvpr/RenZHZM19}&1.50M&&\checkmark&&&29.58&33.09\\
     UMRL~\cite{DBLP:conf/cvpr/YasarlaP19}&984K&&&&& 26.05 &30.15\\
     RESCAN~\cite{DBLP:conf/eccv/LiWLLZ18}&1.04M&\checkmark&\checkmark&&\checkmark&28.90&34.90\\
     \hline
    \end{tabular}
    \label{table1}
    \vspace{-2mm}
\end{table}

\vspace{-1mm}
\subsection{Implementation Details}
\noindent \textbf{1) Datasets.} 
We employ two synthetic deraining datasets (Rain100H~\cite{DBLP:conf/cvpr/YangTFLGY17} and RainCityscape~\cite{DBLP:conf/iccv/HalderLC19}) and some real images~\cite{semi_supervised}.
Rain100H dataset is one of the most commonly used datasets consisting of 1800 paired rain/non-rain images of size $480\times320$ for training and 100 paired images of the same size for testing. PSNR and SSIM of input testing set are 12.05 and 0.3623.
RainCityscape dataset consists of 2875 paired images of size $512\times256$ for training and 100 paired images for testing. PSNR and SSIM of input testing set are 19.71 and 0.7087.
Besides paired images, RainCityscape dataset also provides additional labels for object detection and semantic segmentation, which allow us to evaluate the performance of down-stream tasks.
In our experiment, the model evaluated on the testing set of Rain100H~\cite{DBLP:conf/cvpr/YangTFLGY17} and RainCityscape~\cite{DBLP:conf/iccv/HalderLC19} is trained on their respective training sets. 
\vspace{0.5mm}

\noindent \textbf{2) Deraining Methods.} 
We consider six state-of-the-art deep learning-based deraining models as shown in Table~\ref{table1}.
\vspace{0.5mm}
\noindent \textbf{3) Perturbation Levels.} The adversarial attacks are implemented based on PGD~\cite{DBLP:conf/iclr/MadryMSTV18}. We also set $\epsilon \in E=\left\{1/255, 2/255, 4/255, 8/255\right\}$ regarding to $\ell_{\infty}$-norm, $\alpha = \epsilon/4$, and $T=20$.

\begin{figure*}[h]
    \centering
\begin{subfigure}[b]{1\textwidth}
         \centering
         \includegraphics[width=0.98\linewidth]{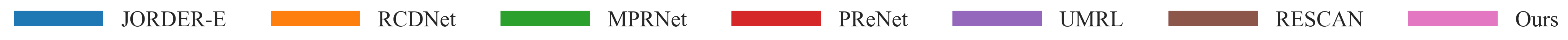}
\end{subfigure}\\
\vspace{-1mm}
\begin{subfigure}[b]{0.16\textwidth}
         \centering
         \includegraphics[width=1\linewidth]{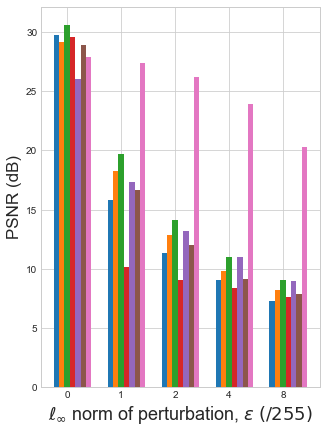}\vspace{-1mm}
         \caption{\small{Rain100H, PSNR}}
\end{subfigure}
\hfill
\begin{subfigure}[b]{0.16\textwidth}
         \centering
         \includegraphics[width=1\linewidth]{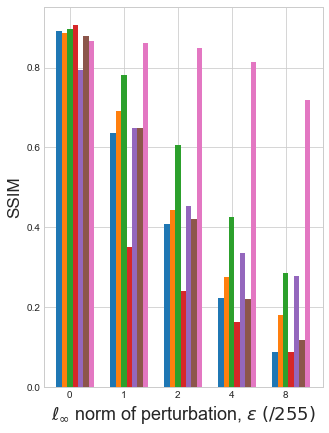}\vspace{-1mm}
         \caption{\small{Rain100H, SSIM}}
\end{subfigure}
\hfill
\begin{subfigure}[b]{0.16\textwidth}
         \centering
         \includegraphics[width=1\linewidth]{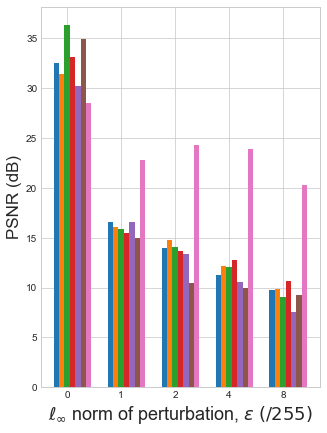}\vspace{-1mm}
         \caption{\small{RC, PSNR}}
\end{subfigure}
\hfill
\begin{subfigure}[b]{0.16\textwidth}
         \centering
         \includegraphics[width=1\linewidth]{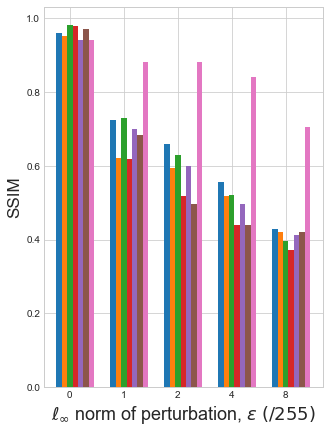}\vspace{-1mm}
         \caption{\small{RC, SSIM}}
\end{subfigure}
\hfill
\begin{subfigure}[b]{0.16\textwidth}
         \centering
         \includegraphics[width=1\linewidth]{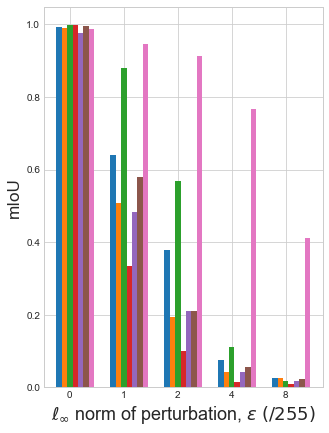}\vspace{-1mm}
         \caption{\small{RC, mIoU}}
\end{subfigure}
\hfill
\begin{subfigure}[b]{0.16\textwidth}
         \centering
         \includegraphics[width=1\linewidth]{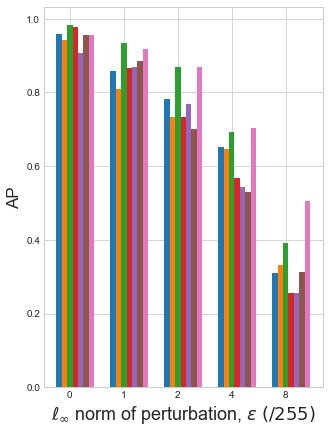}\vspace{-1mm}
         \caption{\small{RC, AP}}
\end{subfigure}\\
\begin{subfigure}[b]{0.16\textwidth}
         \centering
         \includegraphics[width=1\linewidth]{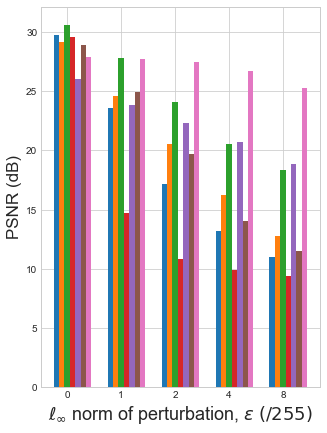}\vspace{-1mm}
         \caption{\small{Rain100H, PSNR}}
\end{subfigure}
\hfill
\begin{subfigure}[b]{0.16\textwidth}
         \centering
         \includegraphics[width=1\linewidth]{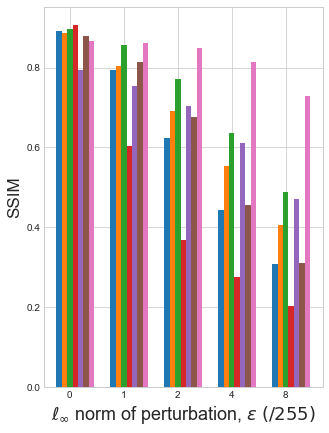}\vspace{-1mm}
         \caption{\small{Rain100H, SSIM}}
\end{subfigure}
\hfill
\begin{subfigure}[b]{0.16\textwidth}
         \centering
         \includegraphics[width=1\linewidth]{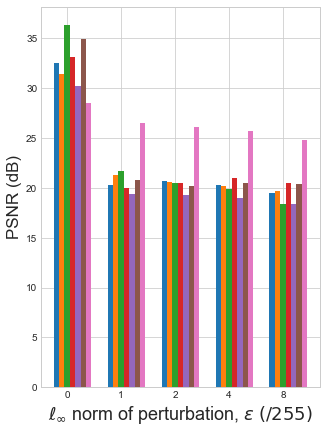}\vspace{-1mm}
         \caption{\small{RC, PSNR}}
\end{subfigure}
\hfill
\begin{subfigure}[b]{0.16\textwidth}
         \centering
         \includegraphics[width=1\linewidth]{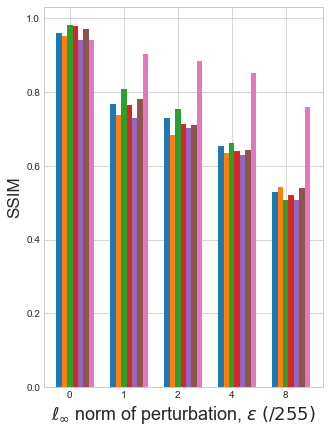}\vspace{-1mm}
         \caption{\small{RC, SSIM}}
\end{subfigure}
\hfill
\begin{subfigure}[b]{0.16\textwidth}
         \centering
         \includegraphics[width=1\linewidth]{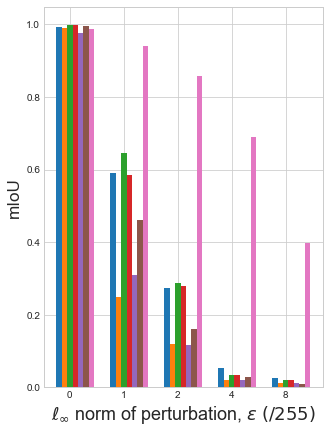}\vspace{-1mm}
         \caption{\small{RC, mIoU}}
\end{subfigure}
\hfill
\begin{subfigure}[b]{0.16\textwidth}
         \centering
         \includegraphics[width=1\linewidth]{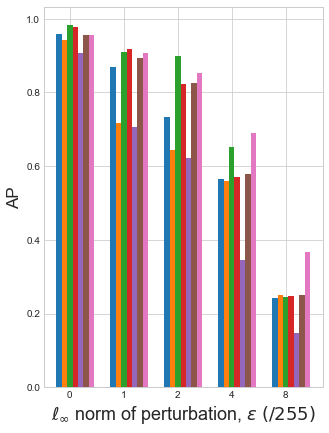}\vspace{-1mm}
         \caption{\small{RC, AP}}
\end{subfigure}
\vspace{-3mm}
    \caption{Adversarial Robustness of six deraining models and our proposed method evaluated by PSNR, SSIM, mIoU of Semantic Segmentation and AP of Pedestrian Detection on the Rain100H~\cite{DBLP:conf/cvpr/YangTFLGY17} and RainCityscape~\cite{DBLP:conf/iccv/HalderLC19} datasets. Subfigures in the first row denote the performance against LMSE attack, and subfigures in the second row denote the LPIPS attack. RC denotes RainCityscape. Note that $\epsilon = 0/255$ denotes the performance of output to input without perturbations.}
    \vspace{-2mm}
    \label{figure1}
\end{figure*}

\begin{figure*}[h]
\centering
\begin{subfigure}[b]{1\textwidth}
         \centering
         \includegraphics[scale=0.205]{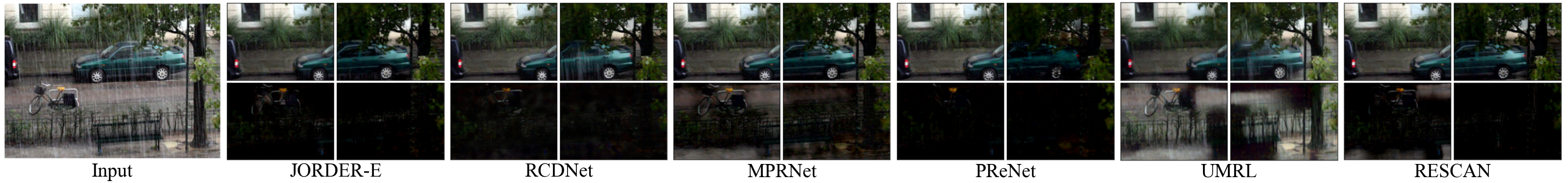}\vspace{-1mm}
         \caption{Deraining Images against adversarial attacks in real images~\cite{semi_supervised}.
The models are trained on Rain100H~\cite{DBLP:conf/cvpr/YangTFLGY17}.}
\end{subfigure}\\
\begin{subfigure}[b]{1\textwidth}
         \centering
         \includegraphics[scale=0.205]{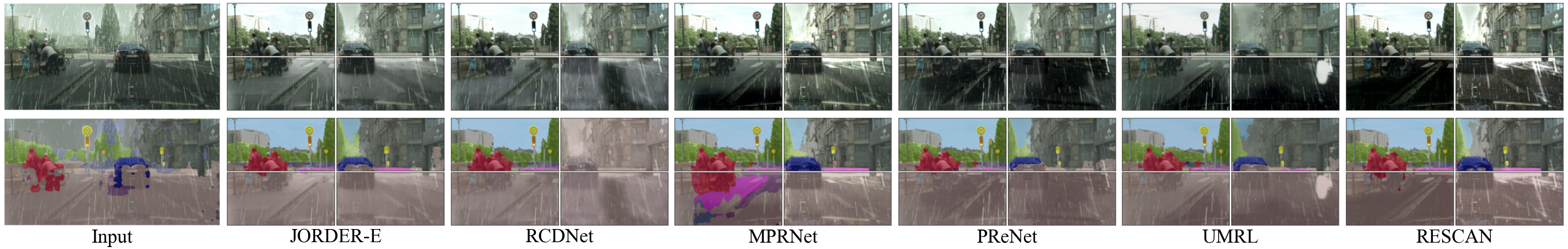}\vspace{-1mm}
         \caption{Deraining Images against adversarial attacks in RainCityscape~\cite{DBLP:conf/iccv/HalderLC19} and its corresponding Semantic Segmentation overlap.}
\end{subfigure}
\vspace{-7mm}
\caption{Visual comparison of the deraining outputs for input with perturbations based on \textbf{LMSE} attack. In each image other than input, patches on the top-left, top-right, bottom-left and bottom-right are corresponding patches from the output to input with perturbation bound $\epsilon=\left\{0/255,1/255,2/255,4/255\right\}$. Best view by zooming in. More visualization results can be found in the supplementary material. }
\vspace{-4mm}
\label{figure2}
\end{figure*}

\begin{table}[t]
    \setlength\tabcolsep{4.5pt}
    \footnotesize
    \centering
    \caption{Adversarial Robustness (mean-Adversarial-Performance) of six deraining models on two datasets. }
    \vspace{-3mm}
    \begin{tabular}{ c || c c | c c c c}
    \hline
    Datasets & \multicolumn{2}{c|}{Rain100H} & \multicolumn{4}{c}{RainCityscape}\\
    \hline
    Metrics & PSNR & SSIM & PSNR & SSIM & mIoU & AP\\
    \hline
    \multicolumn{7}{c}{Restoration (Human  Vision): MSE Loss}\\
    \hline
    JORDER-E & 10.91 & 0.3383 & 12.89 & \textbf{0.5920} & 0.2799 & 0.6511\\
    RCDNet & 12.29 & 0.3744 & \textbf{13.21} & 0.5386 & 0.1924 & 0.6307\\
    MPRNet & \textbf{13.47} & \textbf{0.5253} & 12.76 & 0.5693 & \textbf{0.3945} & \textbf{0.7217}\\
    PReNet & 8.82 & 0.2107 & 13.12 & 0.4864 & 0.1151 & 0.6056\\
    UMRL & 12.94 & 0.4297 & 12.02 & 0.5512 & 0.1886 & 0.6086\\
    RESCAN & 11.40 & 0.3528 & 11.13 & 0.5088 & 0.2168 & 0.6065\\
    \hline
    \multicolumn{7}{c}{Downstream  CV  Tasks (Machine  Vision): LPIPS Loss}\\
    \hline
    JORDER-E & 16.22 & 0.5423 & 20.18 & 0.6692 & 0.2354 & 0.6014\\
    RCDNet & 18.54 & 0.6086 & 20.39 & 0.6498 & 0.1001 & 0.5427\\
    MPRNet & \textbf{22.70} & \textbf{0.6878} & 20.10 & \textbf{0.6815} & \textbf{0.2451} & \textbf{0.6753}\\
    PReNet & 11.22 & 0.3628 & \textbf{20.48} & 0.6598 & 0.2291 & 0.6398\\
    UMRL & 21.44 & 0.6594 & 19.03 & 0.6420 & 0.1140 & 0.4548\\
    RESCAN & 17.54 & 0.5640 & 20.44 & 0.6681 & 0.1652 & 0.6376\\
    \hline
    \end{tabular}
    \label{table2}
    \vspace{-4mm}
\end{table}

\vspace{0.5mm}
\noindent \textbf{4) Results and Remarks.} 
Figure \ref{figure1} and Table \ref{table2} show the robustness results of six deraining models against adversarial attacks.
For the RainCityscape dataset, we also consider the performance of pedestrian detection and semantic segmentation.
From the benchmark results, we can have the following:
\begin{itemize}
    \item 
    \vspace{-1mm}
    Even very small perturbations ($\epsilon = 1 $) can heavily degrade the deraining performance of existing methods.
    \vspace{-0.5mm}
    Larger perturbations $\epsilon$ lead to significantly performance drops in image quality.
    \vspace{-1mm}
    \item On the Rain100H dataset, MPRNet~\cite{DBLP:conf/cvpr/ZamirA0HK0021} is the most robust model in terms of PSNR and SSIM, which might be attributed to its use of attention modules to suppress the effects of adversarial noise at the feature level. 
    \vspace{-1mm}
    \item On the RainCityscape dataset, the gaps of PSNR and SSIM values against adversarial attacks with identical $\epsilon$ are closer for all the models, while MPRNet is most robust for the down-stream tasks in terms of mIoU and AP. 
    \vspace{-1mm}
    \item When attacking with LPIPS, it is observed that perturbations from 0 to 1 have the largest performance drop in terms of PSNR and SSIM while larger perturbations lead to smaller performance gaps. Especially on RainCityscape, perturbations from 1 to 8 bring about the same degradation in terms of PSNR, while SSIM, which considers more structural information, continues to drop. And it's not surprising to find that LMSE attack brings more drop on PSNR and SSIM, while LPIPS attack brings more on mIoU and AP.
    \vspace{-1mm}
    \item It is interesting to note that the deraining results are vulnerable in terms of semantic segmentation performance. The attack with the perturbation ($\epsilon = 2$) has already crashed the segmentation performance.
    \vspace{-1mm}
    \item Existing methods are more robust for object detection than semantic segmentation.
\end{itemize}

\section{Towards Robust Deraining Model}
In this section, we perform a comprehensive analysis on the modules and losses to evaluate their adversarial robustness.

\subsection{Module Analysis}
The state-of-the-art deraining methods usually consist of recurrent blocks, several attention modules, side information injection and larger receptive field. 
We perform an ablation study of these modules based on the baseline of RCDNet~\cite{DBLP:conf/cvpr/WangX0M20} and JOEDER-E~\cite{DBLP:conf/cvpr/YangTFLGY17}.
\vspace{1mm}

\noindent \textbf{1) Recurrent Blocks.}
As there might be different rain streak layers overlapping with each other and it is not easy to remove all streaks in one stage, many deraining models incorporate a recurrent structure to perform rain removal into multiple stages. The process can be formulated as:
\begin{equation}
\begin{aligned}
    O_0 &= O,\\
    [R_t,B_t]&=f^t(O_{t-1},R_{t-1}), 1 \leq t \leq T,\\
    O_t&=O-R_t=B_t, 1 \leq t \leq T,
\end{aligned}
\end{equation}
where $O$ is the input, $T$ is the number of recurrent blocks, $R_t$ is the output rain streaks at the $t$-th stage, $B_t$ is the output background at the $t$-th stage, and $f^t$ is the $t$-th CNN block. In each stage, $R_t$ and $B_t$ are updated, and the final output is $B_T$.
\vspace{1mm}

\begin{figure}[t]
\centering
\includegraphics[scale=0.44]{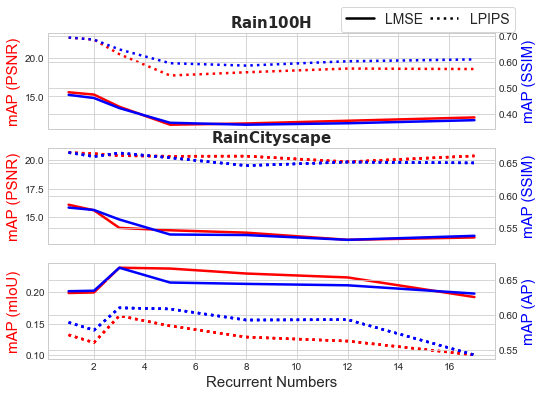}
\vspace{-7mm}
\caption{Recurrent Blocks: Adversarial Robustness for RCDNet with various stage numbers. Red curves correspond to the left y-axis, and blue curves  correspond to the right y-axis.}
\vspace{-2mm}
\label{figure3}
\end{figure}

We examine the adversarial robustness of RCDNet with various stage numbers $\left\{1, 2, 3, 5, 8, 12, 17\right\}$. For a fair comparison, the models with fewer stages have wider feature maps to keep the number of parameters around $3M$. 

\noindent \textbf{Insight}: 
As shown in Fig.~\ref{figure3}, \textit{the models with one or two recurrent blocks are more robust in terms of PSNR and SSIM, while the model with three stages is the most robust considering the mIoU and AP for down-stream tasks and the second most robust in terms of PSNR and SSIM}.

\vspace{1mm}

\begin{table}[t]
    \setlength\tabcolsep{3pt}
    \footnotesize
    \centering
    \caption{Attention Module: Adversarial Robustness of RCDNet with various attention module. Clean represents the performance for input without perturbations in terms of SSIM. }
    \vspace{-3mm}
    \begin{tabular}{ @{\hspace{0.1pt}}c@{\hspace{1pt}} || @{\hspace{1pt}}c@{\hspace{1pt}} | @{\hspace{1pt}}c@{\hspace{2pt}} c @{\hspace{1pt}}| c@{\hspace{1pt}} | @{\hspace{1pt}}c @{\hspace{2pt}} c @{\hspace{2pt}} c @{\hspace{2pt}} c@{\hspace{0.5pt}}}
    \hline
    Datasets & \multicolumn{3}{c|}{Rain100H} & \multicolumn{5}{c}{RainCityscape}\\
    \hline
    Metrics &Clean& PSNR & SSIM &Clean& PSNR & SSIM & mIoU & AP\\
    \hline
    \multicolumn{7}{c}{Restoration (Human  Vision): MSE Loss}\\
    \hline
    RCDNet &0.8867& 12.29 & 0.3744 &0.9517& 13.21 & 0.5386 & 0.1924 & 0.6307\\
    + SE (MUL) &0.8834& \textbf{16.10} & 0.5325 &0.9626& 13.43 & 0.5768 & 0.2874 & 0.6425\\
    + SE (ADD) &\textbf{0.8859}& 15.63 & 0.5246 &\textbf{0.9697}& \textbf{13.83} & \textbf{0.6100} & \textbf{0.3502} & \textbf{0.6672}\\
    + CBAM &0.8777& 14.47 & 0.4930 &0.9553& 13.55 & 0.6039 & 0.2699 & 0.6465\\
    + GC (MUL) &0.8797& 15.83 & 0.5371 &0.9617& 13.75 & 0.5959 & 0.2869 & 0.6666\\
    + GC (ADD) &0.8694& 15.81 & \textbf{0.5595} &0.9644& 10.99 & 0.5648 & 0.2947 & 0.6347\\
    + SK &0.8813& 15.29 & 0.5260 &0.9553& 12.77 & 0.5574 & 0.3478 & 0.6341\\
    \hline
    \multicolumn{7}{c}{Downstream  CV  Tasks (Machine  Vision): LPIPS Loss}\\
    \hline
    RCDNet &0.8867& 18.54 & 0.6086 &0.9517& 20.39 & 0.6498 & 0.1001 & 0.5427\\
    + SE (MUL) &0.8834& 24.00 & 0.7108 &0.9626& 20.23 & 0.6405 & 0.1976 & \textbf{0.6838}\\
    + SE (ADD) &\textbf{0.8859}& 23.68 & 0.7011 &\textbf{0.9697}& 20.44 & 0.6867 & \textbf{0.2461} & 0.6730\\
    + CBAM &0.8777& 22.33 & 0.6552 &0.9553& 19.25 & 0.6783 & 0.1373 & 0.6603\\
    + GC (MUL) &0.8797& \textbf{24.50} & \textbf{0.7176} &0.9617& 20.02 & 0.6574 & 0.2242 & 0.6725\\
    + GC (ADD) &0.8694& 23.81 & 0.7159 &0.9644& 19.43 & 0.6114 & 0.1710 & 0.6516\\
    + SK &0.8813& 23.58 & 0.7048 &0.9553& \textbf{20.48} & \textbf{0.7109} & 0.2389 & 0.5603\\
    \hline
    \end{tabular}
    \label{table3}
    \vspace{-3mm}
\end{table}

\noindent \textbf{2) Attention Module.}
RESCAN~\cite{DBLP:conf/eccv/LiWLLZ18} and MPRNet~\cite{DBLP:conf/cvpr/ZamirA0HK0021} both consist of attention modules to enhance its deraining performance.
We look into their effects in terms of robustness. 
We consider several attention modules based on feature re-calibration in CNN:
\begin{itemize}
    \vspace{-1mm}
    \item \textit{SENet}~\cite{DBLP:conf/cvpr/HuSS18}: 
    Features are first passed through an avg-pooling \emph{squeeze} operator across spatial dimension and then produce the channel-wise activations by the subsequent \emph{excitation} operation. Note that the activations can be added or multiplied to the feature maps.
    \vspace{-1mm}
    \item \textit{GCNet} \cite{DBLP:journals/corr/abs-2012-13375}: GCNet differs from SENet in the \emph{squeeze} operation, and adopts the structures of Non-Local Networks\cite{DBLP:conf/cvpr/0004GGH18} for spatial modeling. 
    \vspace{-1mm}
    \item \textit{CBAM}~\cite{DBLP:conf/eccv/WooPLK18}: CBAM has two sequential sub-modules: channel and spatial. 
    The subsequent module concatenates the output of Max and Avg pooling (across channel), and uses a Conv layer to produce the spatial attention map. 
    \vspace{-1mm}
    \item \textit{SKNet}~\cite{DBLP:conf/cvpr/LiW0019}: "\emph{Selective Kernel}" (SK) block fuses the features in multiple branches with different kernel sizes via softmax attention. Different channel attentions on these branches are performed as well.
\end{itemize}

\noindent \textbf{Insight}: By inserting the attention module into the same location of RCDNet (after each conv block), we find that \textit{all attention modules indeed improve the robustness and SE with the addition is the most robust considering the down-stream tasks; it also performs well in terms of PSNR and SSIM, as can be seen from Table \ref{table3}}.
\vspace{1mm}

\noindent \textbf{3) Side Information.}
A few deraining models require side information during training, \textit{e.g.}, JORDER-E~\cite{DBLP:conf/cvpr/YangTFLGY17} adopts rain mask and density as supervision.
%
%
We look into the effect of this side information on adversarial robustness.

\begin{table}[t]
    \setlength\tabcolsep{3pt}
    \footnotesize
    \centering
    \caption{Side Information: Effect of side information on the adversarial robustness of JORDER-E. Clean represents the performance for input without perturbations in terms of SSIM. }
    \vspace{-3mm}
    \begin{tabular}{ @{\hspace{0.1pt}}c@{\hspace{1pt}} || @{\hspace{2pt}}c@{\hspace{2pt}} | c@{\hspace{2pt}} c @{\hspace{2pt}}| c@{\hspace{2pt}} | @{\hspace{1pt}}c @{\hspace{2pt}} c @{\hspace{2pt}} c @{\hspace{2pt}} c@{\hspace{0.5pt}}}
    \hline
    Datasets & \multicolumn{3}{c|}{Rain100H} & \multicolumn{5}{c}{RainCityscape}\\
    \hline
    Metrics &Clean& PSNR & SSIM &Clean& PSNR & SSIM & mIoU & AP\\
    \hline
    \multicolumn{7}{c}{Restoration (Human  Vision): MSE Loss}\\
    \hline
    JORDER-E &\textbf{0.8921}& 10.91 & 0.3383 &0.9600& 12.89 & 0.5920 & 0.2799 & \textbf{0.6511}\\
    NO Mask &0.8809& \textbf{15.37} & \textbf{0.4991} &\textbf{0.9626}& \textbf{14.89} & \textbf{0.6143} & \textbf{0.2955} & 0.6202\\
    \hline
    \multicolumn{7}{c}{Downstream  CV  Tasks (Machine  Vision): LPIPS Loss}\\
    \hline
    JORDER-E &\textbf{0.8921}& 16.22 & 0.5423 &0.9600& \textbf{20.18} & 0.6692 & 0.2354 & 0.6014\\
    NO Mask &0.8809& \textbf{22.71} & \textbf{0.6947} &\textbf{0.9626}& 20.01 & \textbf{0.6707} & \textbf{0.2574} & \textbf{0.6176}\\
    \hline
    \end{tabular}
    \label{table4}
    \vspace{-4mm}
\end{table}

\noindent \textbf{Insight}: From Table \ref{table4}, JORDER-E without rain mask performs better against adversarial attacks, suggesting that \textit{the model without side information tends to be more robust}.
\vspace{1mm}

\noindent \textbf{4) Receptive Field.}
It is well known that, increasing the receptive field can boost the model’s capacity in handling more complex signal mapping. 
However, it is unclear how different receptive fields affect the robustness of a model. 
Some deraining models adopt multi-path dilated convolution structure to increase its receptive field, which provides a means to analyze the robustness of a model while keeping others unchanged.
JORDER-E~\cite{DBLP:conf/cvpr/YangTFLGY17}’s basic block consists of three paralleled convolution paths with a dilation factor of 1, 2, and 3, respectively. 
By removing paths with larger receptive fields, the effect of receptive fields on adversarial robustness is discussed.

\begin{figure*}[htp]
\centering
\begin{subfigure}[b]{1\textwidth}
         \centering
         \includegraphics[scale=0.186]{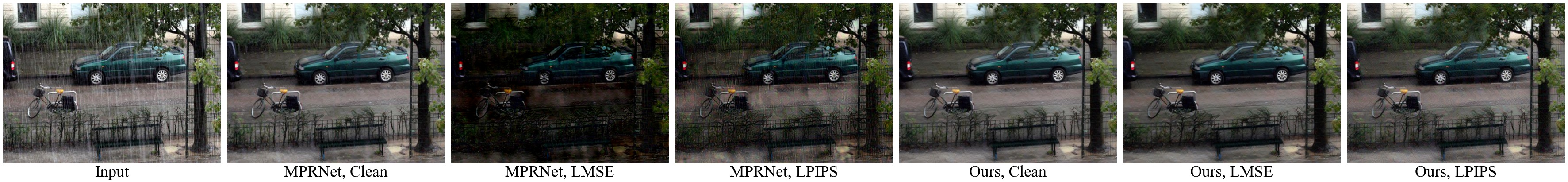}\vspace{-0.5mm}
         \caption{Deraining Images against adversarial attacks in real images~\cite{semi_supervised}.
        The models are trained on Rain100H~\cite{DBLP:conf/cvpr/YangTFLGY17}.}
\end{subfigure}\\
\begin{subfigure}[b]{1\textwidth}
         \centering
         \includegraphics[scale=0.20]{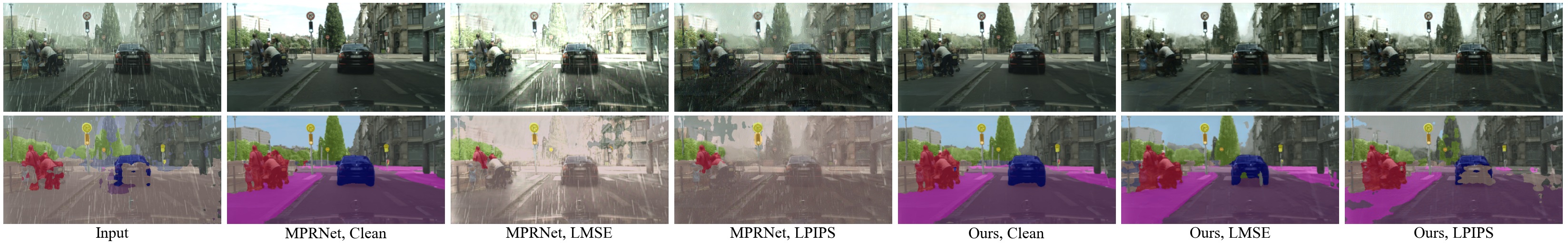}\vspace{-0.5mm}
         \caption{Deraining Images against adversarial attacks in RainCityscape~\cite{DBLP:conf/iccv/HalderLC19} and its corresponding Semantic Segmentation overlap.}
\end{subfigure}
\vspace{-7mm}
\caption{Visual comparison of the derain outputs of RainCityscape~\cite{DBLP:conf/iccv/HalderLC19} test data with adversarial perturbation bound $\epsilon=4/255$. Images annotated with Clean are the output of input without perturbations, and images annotated with LMSE/LPIPS are the output of input with adversarial perturbations. Best view by zooming in. More visualization results can be found in the supplementary material. 
}
\vspace{-3mm}
\label{figure8}
\end{figure*}

\begin{table}[t]
    \setlength\tabcolsep{3pt}
    \footnotesize
    \centering
    \caption{Receptive field: Effect of receptive field on the adversarial robustness of JORDER-E. Clean represents the performance for input without perturbations in terms of SSIM.}
    \vspace{-3mm}
    \begin{tabular}{ @{\hspace{0.1pt}}c@{\hspace{1pt}} || @{\hspace{1pt}}c@{\hspace{1pt}} | @{\hspace{1pt}}c@{\hspace{2pt}} c @{\hspace{2pt}}| c@{\hspace{1pt}} | @{\hspace{1pt}}c @{\hspace{2pt}} c @{\hspace{2pt}} c @{\hspace{2pt}} c@{\hspace{0.5pt}}}
    \hline
    Datasets & \multicolumn{3}{c|}{Rain100H} & \multicolumn{5}{c}{RainCityscape}\\
    \hline
    Metrics &Clean& PSNR & SSIM &Clean& PSNR & SSIM & mIoU & AP\\
    \hline
    \multicolumn{7}{c}{Restoration (Human  Vision): MSE Loss}\\
    \hline
    D = $\left\{1,2,3\right\}$ &0.8921& \textbf{10.91} & \textbf{0.3383} &0.9600& \textbf{12.89} & \textbf{0.5920} & \textbf{0.2799} & \textbf{0.6511}\\
    D = $\left\{1,2\right\}$ &\textbf{0.8969}& 9.32 & 0.2716 &\textbf{0.9622}& 12.60 & 0.5060 & 0.2192 & 0.6154\\
    D = $\left\{1\right\}$ &0.8956& 9.19 & 0.2650 &0.9608& 12.56 & 0.5193 & 0.2268 & 0.6215\\
    \hline
    \multicolumn{7}{c}{Downstream  CV  Tasks (Machine  Vision): LPIPS Loss}\\
    \hline
    D = $\left\{1,2,3\right\}$ &0.8921& \textbf{16.22} & \textbf{0.5423} &0.9600& \textbf{20.18} & \textbf{0.6692} & \textbf{0.2354} & \textbf{0.6014}\\
    D = $\left\{1,2\right\}$ &\textbf{0.8969}& 12.65 & 0.4507 &\textbf{0.9622}& 19.61 & 0.6522 & 0.1405 & 0.5573\\
    D = $\left\{1\right\}$ &0.8956& 12.50 & 0.4426 &0.9608& 20.04 & 0.6607 & 0.1500 & 0.5576\\
    \hline
    \end{tabular}
    \label{table7}
    \vspace{-2mm}
\end{table}

\noindent \textbf{Insight}: Superior adversarial performance can be observed from Table \ref{table7} when a \textit{larger receptive field} is adopted.

\begin{table}[t]
    \setlength\tabcolsep{3pt}
    \footnotesize
    \centering
    \caption{Effect of adversarial loss on the adversarial robustness of RCDNet. Clean represents the performance for input without perturbations in terms of SSIM. }
    \vspace{-3mm}
    \begin{tabular}{ @{\hspace{0.1pt}}c@{\hspace{1pt}} || c | @{\hspace{1pt}}c@{\hspace{2pt}} c @{\hspace{2pt}}| c | @{\hspace{1pt}}c @{\hspace{2pt}} c @{\hspace{2pt}} c @{\hspace{2pt}} c@{\hspace{0.5pt}}}
    \hline
    Datasets & \multicolumn{3}{c|}{Rain100H} & \multicolumn{5}{c}{RainCityscape}\\
    \hline
    Metrics &Clean& PSNR & SSIM &Clean& PSNR & SSIM & mIoU & AP\\
    \hline
    \multicolumn{7}{c}{Restoration (Human  Vision): MSE Loss}\\
    \hline
    RCDNet &\textbf{0.8867}& 12.29 & 0.3744 &\textbf{0.9517}& 13.21 & 0.5386 & 0.1924 & 0.6307\\
    + Adv Loss &0.8584& \textbf{21.85} & \textbf{0.7619} &0.9308& \textbf{17.78} & \textbf{0.6503} & \textbf{0.2855} & \textbf{0.6944}\\
    \hline
    \multicolumn{7}{c}{Downstream  CV  Tasks (Machine  Vision): LPIPS Loss}\\
    \hline
    RCDNet &\textbf{0.8867}& 18.54 & 0.6086 &\textbf{0.9517}& 20.39 & 0.6498 & 0.1001 & 0.5427\\
    + Adv Loss &0.8584& \textbf{25.93} & \textbf{0.8050} &0.9308& \textbf{22.68} & \textbf{0.7230} & \textbf{0.2365} & \textbf{0.6289}\\
    \hline
    \end{tabular}
    \label{table5}
    \vspace{-2mm}
\end{table}

\subsection{Fidelity Loss vs. Adversarial Loss}

Zhang \textit{et al.}~\cite{pmlr-v97-zhang19p} explored the trade-off between accuracy and robustness for classification problems and proposed a new formulation of adversarial loss to defense the adversarial attacks:
\begin{equation}
    L_{\text{adv}}= \lambda \underset{X' \in \mathbb{B}(X,\epsilon)}{\max} \mathcal{L}(f(X | \theta ),f(X' | \theta )),
\end{equation}
where $f(X | \theta )$ is the output vector of learning model, $\mathcal{L}(\cdot, \cdot)$ is a multi-class calibrated loss such as cross-entropy loss, and $\mathbb{B}(X,\epsilon)$ represents the neighborhood of $X: \left\{X': {\left\|X'-X\right\|}_{\infty} \leq \epsilon \right\}$.
The trade-off can be balanced by adjusting the loss weight $\lambda$ in the final combined loss and the perturbation bound $\epsilon$.

For rain removal model, the adversarial loss can be obtained as:
\begin{equation}
    L_{\text{adv}}=\lambda \underset{\delta, {\left\|\delta\right\|}_{\infty} \leq \epsilon}{\max} \left\|f(X+\delta | \theta )-f(X | \theta )\right\|_2.
\end{equation}

\noindent \textbf{Insight}: Table \ref{table5} shows that incoporating  \textit{adversarial loss can improve the adversarial robustness, but it also compromises the performance on input without perturbations}.

\begin{table}[t]
    \setlength\tabcolsep{3pt}
    \footnotesize
    \centering
    \caption{Ablation Study: Robustness of enhanced model compared with MPRNet. Clean represents the performance for input without perturbations in terms of SSIM. AL denotes adversarial loss, AT denotes attention module, RB denotes recurrent blocks and DD denotes diverse dilations. }
    \vspace{-3mm}
    \begin{tabular}{ @{\hspace{0.1pt}}c@{\hspace{1pt}} || @{\hspace{2pt}}c@{\hspace{1pt}} | @{\hspace{1pt}}c@{\hspace{2pt}} c @{\hspace{1pt}}| c@{\hspace{1pt}} | @{\hspace{1pt}}c @{\hspace{2pt}} c @{\hspace{2pt}} c @{\hspace{2pt}} c@{\hspace{0.5pt}}}
    \hline
    Datasets & \multicolumn{3}{c|}{Rain100H} & \multicolumn{5}{c}{RainCityscape}\\
    \hline
    Metrics &Clean& PSNR & SSIM &Clean& PSNR & SSIM & mIoU & AP\\
    \hline
    \multicolumn{7}{c}{Restoration (Human  Vision): MSE Loss}\\
    \hline
    MPRNet &\textbf{0.8974}& 13.47 & 0.5253 &\textbf{0.9811}& 12.76 & 0.5693 & 0.3945 & 0.7217\\
    Ours &0.8670& \textbf{24.46} & \textbf{0.8099} &0.9401& \textbf{22.82} & \textbf{0.8264} & \textbf{0.7592} & \textbf{0.7487}\\
    \hline
    Ours w/o AL &0.8841& 15.35 & 0.5997 &0.9794& 12.67 & 0.5707 & 0.4745 & 0.7328\\
    Ours w/o RB &0.8546& 22.09 & 0.7710 &0.9389& 21.24 & 0.7992 & 0.6512 & 0.7242\\
    Ours w/o DD &0.8685& 22.08 & 0.7782 &0.9466& 22.05 & 0.8023 & 0.7011 & 0.7230\\
    Ours w/o AT &0.8474& 21.41 & 0.7388 &0.9411& 18.75 & 0.7184 & 0.3062 & 0.6552\\
    \hline
    \multicolumn{7}{c}{Downstream  CV  Tasks (Machine  Vision): LPIPS Loss}\\
    \hline
    MPRNet &\textbf{0.8974}& {22.70} & {0.6878} &\textbf{0.9811}& 20.10 & {0.6815} & {0.2451} & {0.6753}\\
    Ours &0.8670& \textbf{26.81} & \textbf{0.8138} &0.9401& \textbf{25.75} & \textbf{0.8484} & \textbf{0.7212} & \textbf{0.7033}\\
    \hline
    Ours w/o AL &0.8841& 24.49 & 0.7110 &0.9794& 21.00 & 0.6811 & 0.3107 & 0.6955\\
    Ours w/o RB &0.8546& 26.58 & 0.8011 &0.9389& 24.84 & 0.8355 & 0.5904 & 0.6814\\
    Ours w/o DD &0.8685& 26.55 & 0.7993 &0.9466& 25.06 & 0.8342 & 0.6352 & 0.6844\\
    Ours w/o AT &0.8474& 25.75 & 0.7689 &0.9411& 22.15 & 0.7530 & 0.2194 & 0.5917\\
    \hline
    \end{tabular}
    \label{table6}
    \vspace{-3mm}
\end{table}

\subsection{Ensemble Modules into a Robust Deraining Model}
Based on the above insights, we build a robust deraining model by integrating  MPRNet, SE with addition, convolution layer with diverse dilations, three recurrent blocks, and adversarial loss. 
Figure~\ref{figure1} and Table~\ref{table6} shows that our model is much more robust than MPRNet.
Figure~\ref{figure8} presents the visual results of different methods.
While MPRNet is already the most robust among all state-of-the-art methods, its restoration results in Figure \ref{figure8}~(a) and (b) as well as the down-stream segmentation results in Figure \ref{figure8}~(c) are still heavily degraded. Benefiting from the robust modules and loss derived from our module analysis, our enhanced model is more robust and obtains visually similar results compared to the ones without adversarial perturbations.

We also conduct an ablation study by subtracting one of the robust modules from our model. We can make the following conclusions from Table \ref{table6}:
\begin{itemize}
    \vspace{-1mm}
    \item  All aforementioned modules contribute to the robustness in our model.
    \vspace{-1mm}
    \item 
    Adversarial loss is the key component to improve robustness at the cost of reducing the performance for input without perturbations a bit.
    \vspace{-1mm}
    \item The attention module can help maintain robustness when the results are evaluated by down-stream tasks.
    \vspace{-2mm}
\end{itemize}

\begin{figure*}[t]
\centering
\begin{subfigure}[t]{0.22\textwidth}
         \centering
         \includegraphics[scale=0.43]{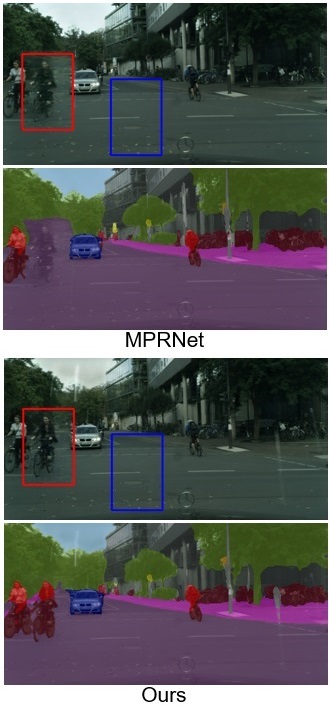}\vspace{-1mm}
         \caption{Object-sensitive Attack}
\end{subfigure}
\begin{subfigure}[t]{0.22\textwidth}
         \centering
         \includegraphics[scale=0.43]{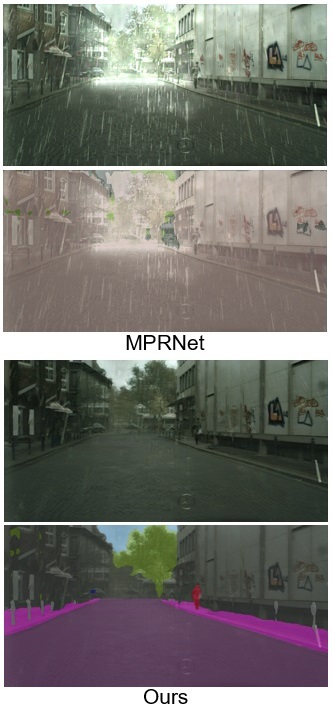}\vspace{-1mm}
         \caption{Partial Attack (Rain Region Attack)}
\end{subfigure}
\begin{subfigure}[t]{0.22\textwidth}
         \centering
         \includegraphics[scale=0.43]{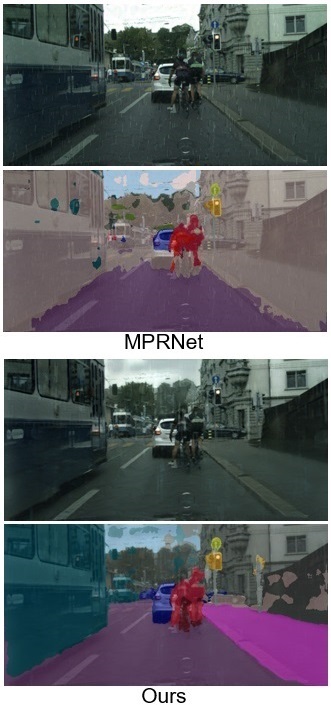}\vspace{-1mm}
         \caption{Unnoticeable Attack for Downstrean Tasks}
\end{subfigure}
\begin{subfigure}[t]{0.22\textwidth}
         \centering
         \includegraphics[scale=0.43]{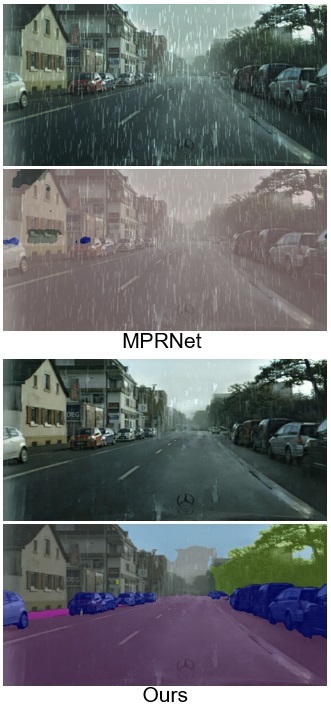}\vspace{-1mm}
         \caption{Input-close  Attack}
\end{subfigure}
\vspace{-3mm}
\caption{Derain Images of \textbf{MPRNet} and \textbf{Ours} against several attacking scenarios and its corresponding semantic segmentation overlap in RainCityscape~\cite{DBLP:conf/iccv/HalderLC19}. Note that the adversarial perturbation bound $\epsilon=4/255$. In (a), the patches with red boxes are the source objects to attack, and patches with blue boxes are the target objects. More visualization results can be found in the supplementary material.
}
\vspace{-4mm}
\label{figure9}
\end{figure*}

\section{Discussion: Advanced Attack Scenarios}
The two basic attacks described in Sec.~\ref{sec:attack_framework} aim to maximize the deviation of the output with the perturbations embedded over the entire input image.
In this section, we investigate other advanced attack scenarios and show that our model can handle them well, especially for the case where human and machine vision applications are considered jointly.

\subsection{Object-sensitive Attack}
In this scenario, the attack aims to compromise the downstream applications on a specific object, such as pedestrians and cars. With such perturbations added to the input, the adversarial attacks minimize the perceptual loss between the source patch $S$ and target patch $T$. Thus, the object at source patch $S$ is more similar to the target patch $T$ from the machine vision perspective. The attack can be formulated as follows:
\begin{align}
    \hat{\delta} &= \underset{\delta, {\left\|\delta\right\|}_{\infty} \leq \epsilon}{\arg\max}- \ell_{pips}(\text{Crop}_Sf(X+\delta|\theta),\text{Crop}_Tf(X|\theta)), \nonumber  \\
    \delta &= {\hat{\delta}} \circ M_S,
\end{align}
where $\text{Crop}_S (\cdot)$ denotes the image-crop operation at the patch $S$, $M_S$ is the  binary mask specifying the source patch $S$, and $\circ$ represents the element-wise multiplication. We perform the object-sensitive attack for both MPRNet and our proposed method, and the results are shown in Figure~\ref{figure9} (a). In the output of MPRNet, one person riding a bicycles in the source patch ({\color{red}red box}) was mis-classified as the ground in the target patch ({\color{blue}blue box}) by the semantic segmentation approach SSeg~\cite{DBLP:conf/cvpr/0001SRSNTC19}, while the output of our model survived the attack. 

\subsection{Partial Attack (Rain Region Attack)}
We also consider the case that perturbations are added to part of the input image. The attack can be formulated  as follows:
\begin{align}
    \hat{\delta} &= \underset{\delta, {\left\|\delta\right\|}_{\infty} \leq \epsilon}{\arg\max}
    D\left( f(X|\theta), f(X+\delta \circ M |\theta))\right) , \\
    \delta &= {\hat{\delta}} \circ M,
\end{align}
where $M$ is the  binary mask specifying the attack region.
Figure~\ref{figure9} (b) shows the results of MPRNet and our method attacked by adding the perturbations to the rain region. The results show that our method is more robust than MPRNet under such an attack.

\subsection{Unnoticeable Attack for Down-stream Tasks}
We also examine the attack of which the output degradation is unnoticeable but can much affect the performance of downstream tasks, \textit{i.e.} benefiting human vision but degenerating machine vision.
Our objective is to maximize the perceptual loss deviation and minimize the MSE loss deviation in the output, which can be described as follows:
\begin{align}
    \delta = \underset{\delta, {\left\|\delta\right\|}_{\infty} \leq \epsilon}{\arg\max}
    \ell_{pips}(f(X+\delta|\theta),f(X|\theta)) \nonumber \\
    - \lambda \left\|f(X+\delta|\theta)-f(X|\theta)\right\|_2.
\end{align}
where $\lambda$ balances the trade-off between human vision and machine vision deterioration. From the example in Figure~\ref{figure9}~(c), both results of MPRNet and the proposed method are visually pleasing, but our model is much more robust against this attack in the downstream segmentation task.

\subsection{Input-close Attack}
The goal of input-close attack is to minimize the MSE loss between the input and output. 
That is to say, we hope the attacked model totally fails to remove the rain streaks of the input:
\begin{align}
    \delta = \underset{\delta, {\left\|\delta\right\|}_{\infty} \leq \epsilon}{\arg\max}
    -\left\|f(X+\delta|\theta)-X\right\|_2.
\end{align}
Figure~\ref{figure9} (d) shows that our model can handle this attack.

\section{Conclusion}
In this paper, we conducted the first comprehensive study on the robustness of deep learning-based rain streak removal methods against adversarial	attacks.
Based on the analysis, we proposed a more robust deraining method by integrating effective modules.
We also evaluated several types of adversarial attacks that are specific to the properties of rain/deraining problems and exploited the properties or requirements of both human and machine vision tasks.

\textbf{Acknowledgement.} This work was done at Rapid-Rich Object Search (ROSE) Lab, Nanyang Technological University. This research is supported in part by the NTU-PKU Joint Research Institute (a collaboration between the Nanyang Technological University and Peking University that is sponsored by a donation from the Ng Teng Fong Charitable Foundation).
\newpage
\clearpage
{\small
\bibliographystyle{ieee_fullname}
\bibliography{egbib}
}

\end{document}